\documentclass[10pt,twocolumn,letterpaper]{article}

\usepackage{iccv}
\usepackage{times}
\usepackage{epsfig}
\usepackage{graphicx}
\usepackage{amsmath}
\usepackage{amssymb}

\usepackage[ruled]{algorithm2e}
\usepackage{multirow}
\usepackage{float}
\usepackage{amsfonts}
\usepackage{bm}
\usepackage{array}
\usepackage{threeparttable}
\usepackage{color}
\usepackage{subfigure}
\usepackage{epstopdf}
\usepackage[]{booktabs}
\usepackage{multirow}

\newcommand{\beforefigcaption}{\vspace{0mm}}
\newcommand{\afterfigcaption}{\vspace{0mm}}

\newcommand{\beforetab}{\vspace{0mm}}
\newcommand{\aftertab}{\vspace{0mm}}
\newcommand{\beforesection}{\vspace{0mm}}
\newcommand{\aftersection}{\vspace{0mm}}
\newcommand{\beforesubsection}{\vspace{0mm}}
\newcommand{\aftersubsection}{\vspace{0mm}}

\usepackage{etoolbox}
\newcommand{\zerodisplayskips}{%
  \setlength{\abovedisplayskip}{3pt}%
  \setlength{\belowdisplayskip}{3pt}%
  \setlength{\abovedisplayshortskip}{0pt}%
  \setlength{\belowdisplayshortskip}{0pt}}
\appto{\normalsize}{\zerodisplayskips}
\appto{\small}{\zerodisplayskips}
\appto{\footnotesize}{\zerodisplayskips}

\makeatletter
\newcommand{\thickhline}{%
    \noalign {\ifnum 0=`}\fi \hrule height 1pt
    \futurelet \reserved@a \@xhline
}
\makeatother

\usepackage[pagebackref=true,breaklinks=true,letterpaper=true,colorlinks,bookmarks=false]{hyperref}

\iccvfinalcopy 

\def\iccvPaperID{2160} 

\ificcvfinal\fi

\begin{document}

\title{DenseRaC: Joint 3D Pose and Shape Estimation by Dense Render-and-Compare}

\author{Yuanlu Xu$^{1,2}$\quad\quad\quad Song-Chun Zhu$^{2}$\quad\quad\quad Tony Tung$^1$\\
$^1$Facebook Reality Labs, Sausalito, USA\quad\quad $^2$University of California, Los Angeles, USA\\
{\tt\small merayxu@gmail.com, sczhu@stat.ucla.edu, tony.tung@fb.com}
}

\maketitle
\ificcvfinal\thispagestyle{empty}\fi

\begin{abstract}

We present DenseRaC, a novel end-to-end framework for jointly estimating 3D human pose and body shape from a monocular RGB image. Our two-step framework takes the body pixel-to-surface correspondence map (\ie, IUV map) as proxy representation and then performs estimation of parameterized human pose and shape.
Specifically, given an estimated IUV map, we develop a deep neural network optimizing 3D body reconstruction losses and further integrating a render-and-compare scheme to minimize differences between the input and the rendered output, \ie, dense body landmarks, body part masks, and adversarial priors.
To boost learning, we further construct a large-scale synthetic dataset (MOCA) utilizing web-crawled Mocap sequences, 3D scans and animations. The generated data covers diversified camera views, human actions and body shapes, and is paired with full ground truth. Our model jointly learns to represent the 3D human body from hybrid datasets, mitigating the problem of unpaired training data. Our experiments show that DenseRaC obtains superior performance against state of the art on public benchmarks of various human-related tasks. 

\end{abstract}

\beforesection
\section{Introduction}
\aftersection

Though much progress has been made in human pose estimation, body segmentation and action recognition, it remains underexplored to leverage such estimations into the 3D world, due to the difficulty in data acquisition, ambiguities from monocular inputs and nuisances in natural images (\eg, illumination, occlusion, texture). Existing learning-based methods~\cite{kanazawa2018hmr,NeuralBodyFit18,varol18_bodynet} heavily rely on sparse 2D/3D landmarks (\ie, skeleton joints), body part masks or silhouettes. However, it is ambiguous to recover 3D human pose and body shape from such limited information.

\begin{figure}[ptb]
\centering
\includegraphics[width=\linewidth]{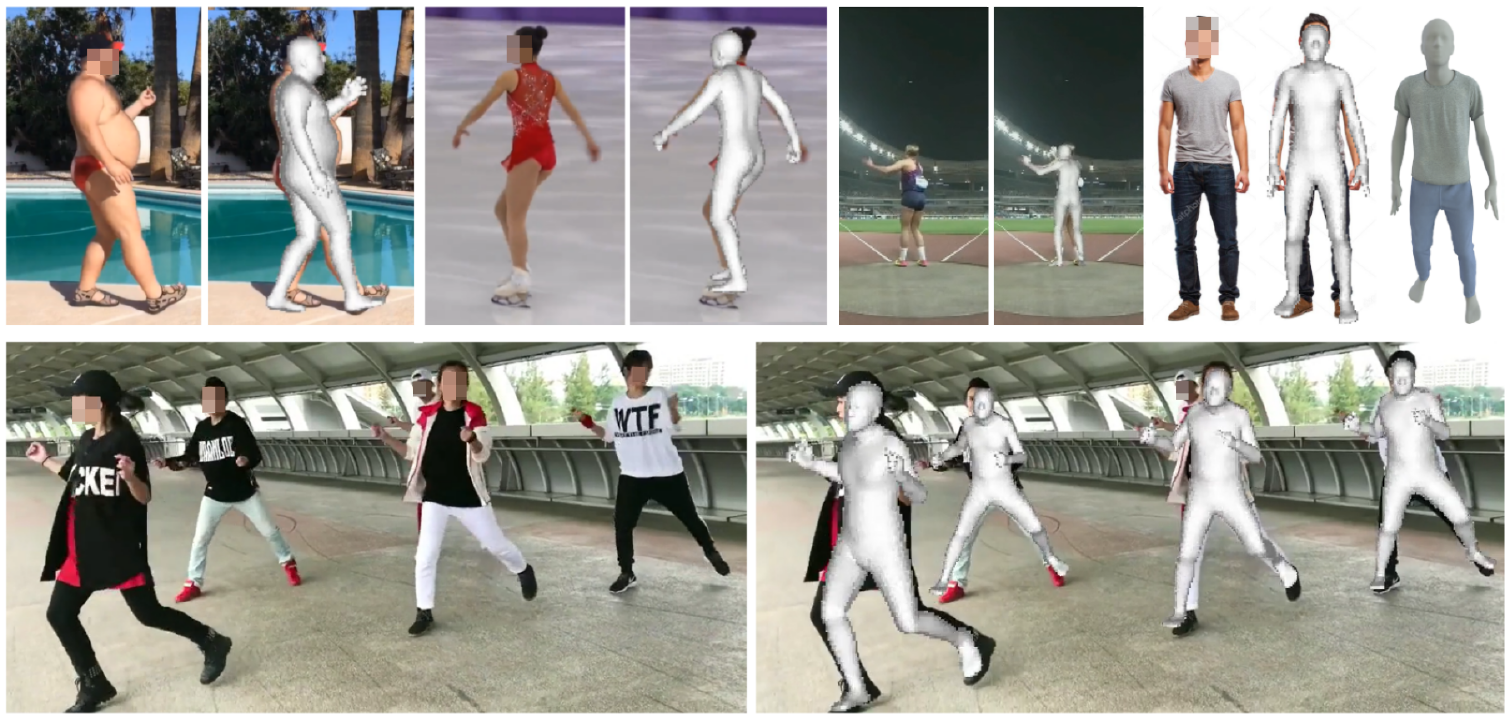}
\beforefigcaption
\caption[DenseRaC estimates 3D human poses and body shapes given people-in-the-wild images. The proposed framework handles scenarios with multiple people, all genders, and various clothing in real time. Here, we show results on Internet images]{DenseRaC estimates 3D human poses and body shapes given people-in-the-wild images. The proposed framework handles scenarios with multiple people, all genders, and various clothing in real time. Here, we show results on Internet images~\cite{internetimages}.}
\afterfigcaption
\label{fig:intro}
\end{figure}

\begin{figure*}[ptb]
\centering
\includegraphics[width=\textwidth]{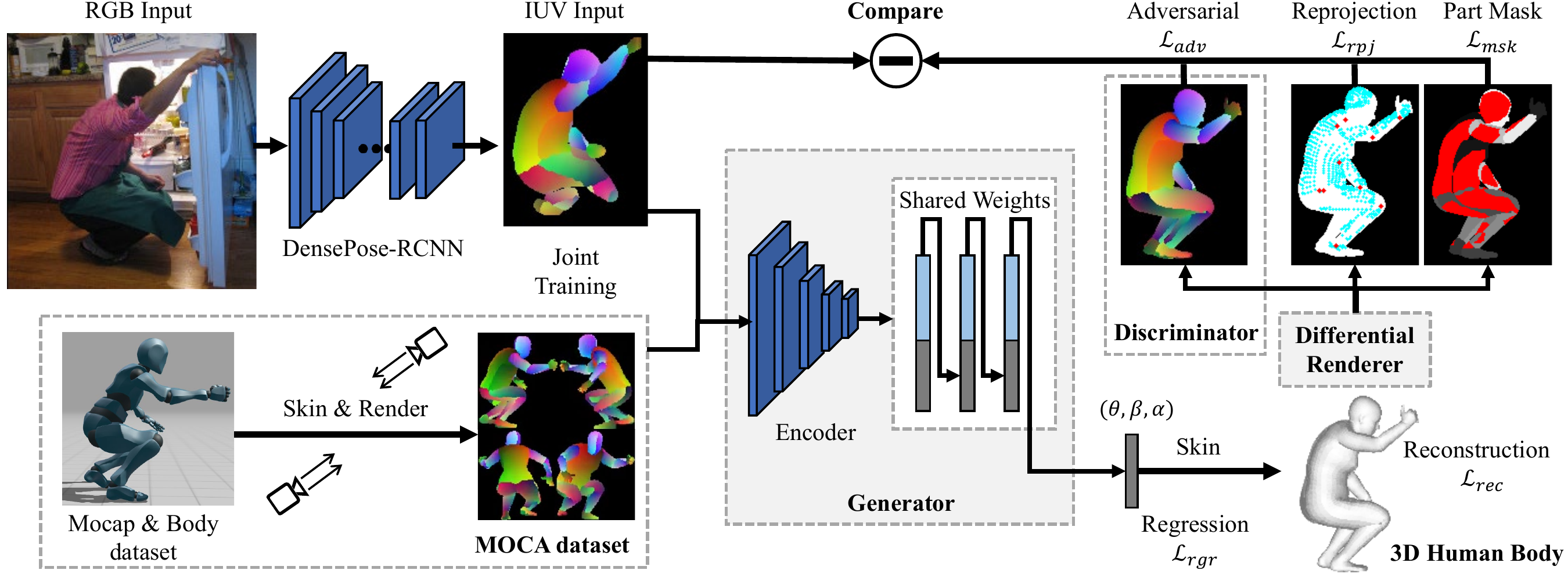}
\beforefigcaption
\caption{Illustration of DenseRaC. Our two-step framework uses pixel-to-surface correspondences of human body as the intermediate representation, fed with data sources either from estimations on realistic images through DensePose-RCNN or rendered images on synthetic 3D humans. Given IUV images, we develop a deep neural network conducting parametric pose and shape regression and a differentiable renderer performing render-and-compare. The proposed framework optimizes losses of 3D reconstruction and discrepancies between inputs and rendered outputs by end-to-end learning.}
\afterfigcaption
\label{fig:model}
\end{figure*}

In this paper we propose DenseRaC, a new framework for 3D human pose and body shape estimation from monocular RGB image, as illustrated in Fig.~\ref{fig:model}:

\noindent {\small \textbullet} \, The task is solved in a two-step framework, first by estimating pixel-to-surface correspondences (\ie, IUV images) from the RGB inputs, and then by leveraging the estimated IUV images into 3D human pose and body shape. 

\noindent {\small \textbullet} \, A parametric human pose and body shape representation is integrated into the forward pass and backward propagation, inspired by recent work~\cite{kanazawa2018hmr,NeuralBodyFit18}.

\noindent {\small \textbullet} \, An IUV image based dense render-and-compare scheme is incorporated into the framework. We minimize 3D reconstruction errors as well as discrepancies between inputs and rendered images from estimated outputs.

We learn the proposed model with both unpaired and paired data, compatible with different levels of supervisions. The end-to-end training minimizes multiple losses defined upon human pose and body shape jointly, including parameter regression, 3D reconstruction, landmark reprojection, body part segmentation, as well as adversarial loss on impossible configurations (see Sec.~\ref{sec:randc}).

To boost learning, we further construct a large scale synthetic dataset covering diversified human poses and body shapes. The synthetic data is generated using web-crawled 3D animations and scanned all-gender body shapes for human studies, and rendered from various camera views (see Sec.~\ref{sec:dataset}). Learning from synthetic data mitigates the problem of unpaired, partial paired, or inaccurately annotated training data in popular public people-in-the-wild and Mocap benchmarks, as well as improves the model robustness against varied camera views and occlusions.

In our experiments, we evaluate DenseRaC on three tasks: 3D pose estimation, semantic body segmentation and 3D body reconstruction. Qualitative and quantitative experimental results show DenseRaC outperforms existing methods on both public benchmarks and the newly proposed synthetic dataset (see Sec.~\ref{sec:experiments}).

To the best of our knowledge, this is the first end-to-end framework introducing a pixel-to-surface correspondence map as the intermediate representation and a corresponding dense render-and-compare scheme for learning 3D human pose and body shapes. We believe DenseRaC shows a great potential for numerous real-world applications in surveillance, entertainment, AR/VR, etc. Some featured results are shown in Fig.~\ref{fig:intro}.

\beforesection
\section{Related Work}
\aftersection

The proposed method is mainly related to researches in three fields.

\textbf{Monocular 3D pose estimation} is a longstanding problem in computer vision. Current approaches train deep networks from large-scale training sets to regress 3D human joint transformations~\cite{ionescu2014human3,li2015maximum}. Deep neural net architectures enable direct body location with pose prediction, which is an advantage compared to traditional model-based methods that require good initialization~\cite{bogo2016keep,lassner2017unite}. Several methods predict 3D pose directly given monocular data~\cite{tekin2016structured,pavlakos2017volumetric,Sun2017,Nie2017,diogoCVPR182d3dml,mirECCV18temp3dpose,umarECCV1825Dhtmp3dpose,helgeECCV18GeoAware}. On the other hand, many approaches lift 2D human poses~\cite{fang2017rmpe,cao2017realtime}, used as intermediate representation, and learn a model for 2D-3D pose space mapping~\cite{yasin2016dual,zhou2016sparseness,zhou2017towards,martinez2017simple,Fang2018}. State of the art in this track obtains fascinating performance on popular benchmarks limited to laboratory instrumented environments, and yet shows unsatisfactory results on in-the-wild images. Another common issue is that most existing methods do not incorporate a physically plausible human skeleton model and lack constraints on the estimated results, which results in extra post-processings for graphics related applications.

\textbf{3D human body reconstruction} aims at recovering full 3D meshes of the human body from single RGB images or video sequences, rather than major 3D skeleton joints. For example, Zuffi \etal~\cite{zuffi2015stitched} integrated both realistic body model and part-based graphical models~\cite{xu2013reid,XuFashionCVPR2018,xu2018caog} for jointly emphasizing graphics-like models of human body shape and part-based human pose inference.
In~\cite{loper2015smpl,bogo2016keep,lassner2017unite,tung2017self}, a skinned body model (SMPL) is used to formulate body shape as a linear function of deformation basis (\ie, with blend shapes). In~\cite{TanBC17,pavlakos2018ps,kanazawa2018hmr,NeuralBodyFit18}, SMPL is considered as the parametric representation of 3D human body and DNNs are developed to estimate such parameters end-to-end. Guler \etal~\cite{guler2017densereg,DensePose2018} build a FCN for human shape estimation by learning dense image-to-template correspondences. Other work~\cite{HSNetsl16,varol18_bodynet,ShapeFMask18Arxiv} focuses on reconstructing 3D body shapes using RGB or RGBD images and not directly estimates 3D human pose and body shapes.
These approaches are also suitable for multiple-view video capture setup~\cite{3Dvideo12,tungICCV09}.
In this paper, we use a SMPL variant as the parametric representation of 3D human body and further develop a pixel-to-surface dense correspondence based render-and-compare framework.

\noindent\textbf{Learning from synthetic humans}. Modeling 3D humans in arbitrary scenes requires representative training sets.
A number of previous work has considered automatically generating data for assisting 3D models, \eg, upper body~\cite{paul2003fast}, full-body silhouettes~\cite{agarwal2006recovering}. \cite{hattori2015learning} artificially renders pedestrians in a scene while leveraging camera parameters and geometrical layout, and further trains a scene-specific pedestrian detector.
In~\cite{pishchulin2012articulated}, real 2D pose samples are reshaped by adding small perturbations, and augmented with different backgrounds. Rogez \etal~\cite{rogez2016mocap}, for a given 3D pose, combines local image patches from several images with kinematic constraints to create a new synthetic image. Rahmani \etal~\cite{rahmani20163d} fits synthetic 3D human models to Mocap skeletons and renders human poses from numerous virtual viewpoints. Varol \etal~\cite{varol2017learning} also generate a synthetic human body dataset with random factors (\eg, pose, shape, texture, background, \etc). 
These datasets cannot solely serve to train models generalized to real data, due the gap between synthesized and realistic images. In this paper, we propose to use pixel-to-surface correspondence maps to bridge the gap. The joint training on hybrid datasets is proved to be effective in improving performance on realistic data. To our best knowledge, we are the first to address joint human pose and body shape estimation using such training modalities.

\beforesection
\section{DenseRaC Framework}
\aftersection

\begin{figure}[ptb]
\centering
\includegraphics[width=\linewidth]{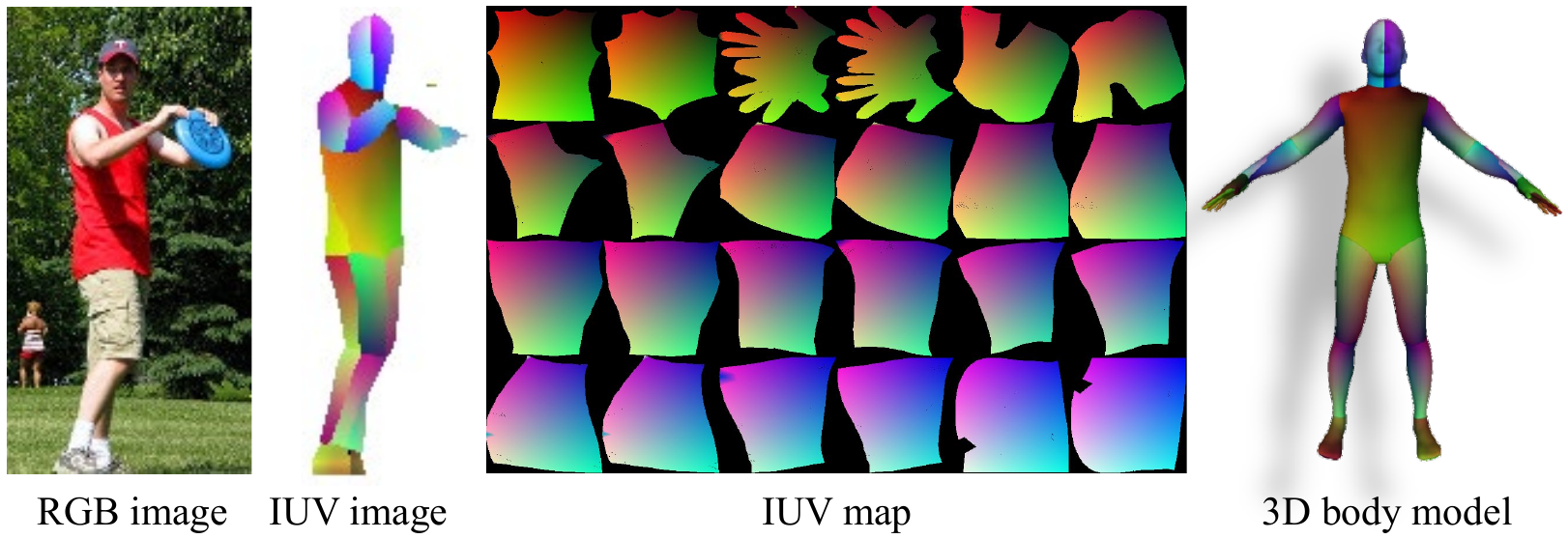}
\beforefigcaption
\caption{Illustration of mapping from pixel to 3D surface. Our framework estimates an IUV image and dense 3D landmarks from an RGB input, whose pixels refer to 3D points on the body model.}
\afterfigcaption
\label{fig:densepose}
\end{figure}

As illustrated in Fig.~\ref{fig:model}, the proposed framework estimates 3D human poses and body shapes in two steps: first obtaining pixel-to-surface correspondences (\ie, IUV images) and then leveraging the intermediate results IUV images into 3D surfaces. There are two sources of IUV inputs: i) estimations from RGB inputs using a pre-trained DensePose-RCNN~\cite{DensePose2018}, and ii) rendered IUV images from synthetic data.

Our framework employs a compact and expressive 3D human body model, which is parameterized by 3D human pose $\theta \in \mathbb{R}^{58 \times 3}$, body shape $\beta \in \mathbb{R}^{50}$, instead of directly estimating 3D point clouds, voxels or depth maps. The 3D human pose is represented as a tree structure, with 58 relative 3D rotations between parent and child joints while the body shape is represented by 50 shape coefficients, as elaborated in Sec.~\ref{sec:bodymodel}.

\beforesubsection
\subsection{Network Architecture} \label{sec:network}
\aftersubsection

Given IUV inputs, we design a network architecture consisting of three modules:

\noindent {\small \textbullet} \, A generator with a back-boned base network (\ie, ResNet-50~\cite{he2016deep}) to extract expressive feature maps and a regressor which takes the stretched feature maps (\ie, 2048D feature vector) from the base network as inputs and estimates 3D human body parameters $[\theta, \beta]$ and camera parameters $\alpha \in \mathbb{R}^3$ (\ie, 227D concatenated vector). The camera model is assumed to be an orthographic projection, parameterized by scale factor $f$ and camera axis $(x, y)$. 
The regressor is composed of 3 fully connected layers with 1024 nodes each. Inspired by~\cite{kanazawa2018hmr}, we consider the regressor to model an iterative update $\Delta_{\theta,\beta,\alpha}$ to the final output, starting from the parameter mean $[\bar{\theta}, \bar{\beta}, \bar{\alpha}]$. The weights are shared across all three layers, simulating the recursive tree structure within 3D human pose.

\noindent {\small \textbullet} \, A differentiable renderer creates 2D projections of the reconstructed 3D human body mesh, using the estimated camera parameters (see Sec.~\ref{sec:randc}). We implement a differential rasterizer which creates an IUV image suitable for gradient flow. Following a render-and-compare scheme, we define three losses to measure and minimize the differences between the input IUV image and the rendered IUV image from our model output.

\noindent {\small \textbullet} \, A discriminator to constrain impossible configurations for unpaired data. We design two shallow networks with two fully connected layers as a discriminator. One is used for discriminating 3D human poses and the other one for body shapes. The number of nodes for pose and shape in sub-networks are set to 512 and 64, respectively. 

\beforesubsection
\subsection{IUV as Proxy Representation} \label{sec:IUVrep}
\aftersubsection

As illustrated in Fig.~\ref{fig:densepose}, we utilize the IUV image as a proxy representation. An IUV map, similarly to UV map in graphics, defines pixel-to-surface correspondences (one-to-one), from 2D image to 3D surface mesh. Each pixel of an IUV image refers to a body part index $I$, and $(U,V)$ coordinates that map to a unique point on the body model surface (see Sec.~\ref{sec:bodymodel}).

As also discussed in~\cite{NeuralBodyFit18}, RGB input contains much more information of the human target than 2D joints, silhouettes, or body part masks that are traditionally used as proxy representation. However information such as appearance, illumination or clothing may not be relevant for inferring the 3D geometry, and even overfits the network to nuisance factors. Similar to ~\cite{NeuralBodyFit18}, we also observe that explicit body part representations are more useful for the task of 3D human pose and body shape estimation, compared to RGB images and plain silhouettes. Better part segmentation produces better 3D reconstruction accuracy, while providing full spatial coverage of the person (compared to joint heatmaps).
While further increasing the number of segmentation parts above a certain threshold (12) only incrementally leverage 3D pose prediction accuracy, it nevertheless greatly improves body shape estimation (see Sec.~\ref{sec:experiments}).
We argue that prior work only estimates average body shape.

Note we further use two sources of IUV images as inputs, \ie, IUV images from realistic images estimated from~\cite{DensePose2018} and IUV images from virtual humans synthesized by our renderer (see Sec.~\ref{sec:randc}). The IUV estimation could be obtained by other off-the-shelf models or two-stage/end-to-end training.
Both inputs go through our neural network model and are used to estimate 3D human pose and body shape parameters.
Thus, there are several benefits for using IUV image representation: i) improving robustness against nuisances of light and texture in natural images, ii) providing richer geometry information on 3D human body (by including body part masks and dense landmarks), iii) unifying realistic and synthetic data for joint learning. 

\beforesubsection
\subsection{Dense Render-and-Compare} \label{sec:randc}
\aftersubsection

In this paper, 3D human pose and body shape are represented compactly by a parametric model (see Sec.~\ref{sec:bodymodel}).
Parametrized 3D human body is inferred and fit to the input image, given also camera parameters.
Human body surface is represented as a 3D triangular mesh, and body posing is obtained by standard linear blend skinning.
To fully compare a reconstructed 3D human body to a 2D observation of it, we integrate a differentiable renderer, \ie, a computer graphics technique that creates a 2D image from a 3D object using differentiable operations~\cite{loper2014opendr,kato2018renderer}, and develop an end-to-end weakly-supervised training scheme.

Rendering consists of projecting 3D vertices of a mesh onto a 2D image plane and rasterizing it (\ie, sampling the faces).
3D-to-2D projection is obtained by a combination of differentiable transformations~\cite{marschner2015}.
Rasterization is a discrete operation that requires gradient definition to allow back-propagation in a neural network. 
In~\cite{loper2014opendr}, the authors approximate derivatives at occlusion boundaries which are discontinuous, while colors are interpolated between vertices (\ie, there is no differentiation with respect to texture).
In~\cite{kato2018renderer}, the authors obtain approximate gradients by blurring image to avoid sudden pixel color change. This produces non-zero gradients and enables gradient-flow between pixel (color) value to vertex position.
However, lighting and material properties in natural images are complex to model and integrate into neural networks.

On the contrary, our IUV representation is invariant to background, lighting conditions and surface texture like clothing (see Sec.~\ref{sec:IUVrep}).
In addition, UV values on each body part I are continuous with respect to neighbor pixels (see Fig.~\ref{fig:densepose}).
This actually allows to naturally compute gradients on mesh surface and at boundaries and back-propagate them through network layers.

\begin{figure}[ptb]
\centering
\includegraphics[width=\linewidth]{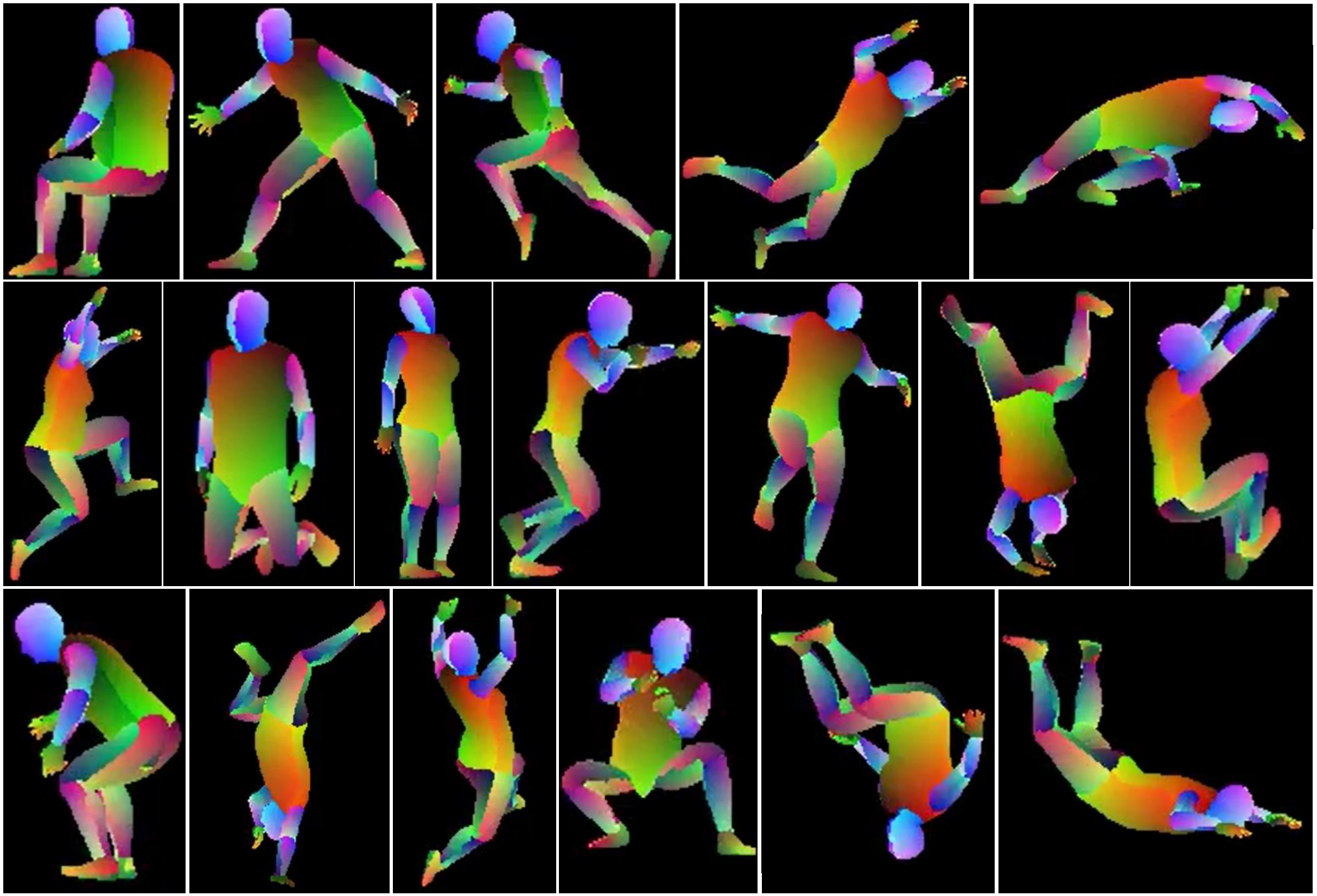}
\beforefigcaption
\caption{IUV images from MOCA generated by rasterizing 3D bodies obtained with 3D poses from Mixamo and body shapes from CAESAR. MOCA contains 2M+ images with fully paired ground truth.}
\afterfigcaption
\label{fig:moca}
\end{figure}

Our renderer creates IUV image comparable to the generated output of~\cite{DensePose2018} (see Fig.~\ref{fig:moca}).
Self-occlusion is handled by depth buffering. Our rasterizer draws only the surface faces closest to the camera (and facing it) at each pixel.
During back propagation, we only pass gradient flows of pixels corresponding to visible regions.

Different from~\cite{tung2017self,Kundu18RenderCompare} where render-and-compare losses are computed upon silhouettes and 2D depth maps, we compute dense render-and-compare losses $\mathcal{L}_{rac}$ using IUV values between ground-truth IUV images and rendered ones (see Sec.~\ref{sec:lossterms}).
The differentiable renderer (including IUV rasterizer) and losses are implemented with differentiable operations using a neural net framework with automatic differentiation~\cite{Dayan95AlySyn,tung2017self,Kundu18RenderCompare}.

\beforesubsection
\subsection{Loss Terms} \label{sec:lossterms}
\aftersubsection

Our model integrates a dense render-and-compare module with corresponding loss computations in the backward propagation, hence leveraging previous methods~\cite{pavlakos2018ps,kanazawa2018hmr,NeuralBodyFit18,varol18_bodynet}.
The loss function is defined as
\begin{equation}\small
\mathcal{L} = \mathcal{L}_{rac} + \mathbf{1} \mathcal{L}_{rec} + \mathbf{1} \mathcal{L}_{rgr},
\end{equation}
where $\mathbf{1}$ indicates the existence of such annotation, $\mathcal{L}_{rac}$, $\mathcal{L}_{rec}$ and $\mathcal{L}_{rgr}$ denote render-and-compare loss, 3D reconstruction loss and parameter regression loss, respectively.

\noindent {\small \textbullet} \, \textbf{Render-and-Compare Loss} $\mathcal{L}_{rac}$ is evaluated under three measurements, that is,
\begin{equation}\small
\mathcal{L}_{rac} = \mathcal{L}_{rpj} + \mathcal{L}_{msk} + \mathcal{L}_{adv},
\end{equation}
where $\mathcal{L}_{rpj}$, $\mathcal{L}_{msk}$ and $\mathcal{L}_{adv}$ denote landmark reprojection loss, part mask loss and adversarial loss, respectively.

\textbf{Landmark Reprojection Loss} $\mathcal{L}_{rpj}$ measures displacement between ground truth and estimated dense 2D landmarks: 
\begin{equation}\small
\mathcal{L}_{rpj} = \sum\nolimits_{i}^{N} \mathbf{1}_i \|p_i - \hat{p}_i \|_1,
\end{equation}
where $\mathbf{1}_i$ indicates the visibility (1 if visible, 0 otherwise) for $i$-th 2D landmark ($N$ in total),
$p_i \in \mathbb{R}^{2}$ and $\hat{p}_i \in \mathbb{R}^{2}$ represent $i$-th 2D landmark from ground truth and 3D mesh reprojection, respectively.
To correctly localize the landmarks from ground truth (\ie, IUV image estimated from DensePose~\cite{DensePose2018}), we formulate this problem as a point-to-point greedy match and solve the correspondence problem by k-Nearest Neighbor (k-NN) search. 
Specifically, we first create a k-D tree for IUV values of 3D body mesh vertices.
For any input IUV image, we search for 1-NN of each visible pixel and obtain a matched pair with the closest 3D body mesh vertex within a distance threshold $\tau$.
Empirically, $\tau \in [0.01, 0.1]$ yields 100-300 matching pairs considered as near-optimal one-to-one 2D/3D dense landmarks correspondences.
This serves as a weakly-supervised scaffold to densely fit 3D human body to the re-projected 2D image. Note the matching is computed offline and serves as a pre-processing step on IUV inputs, as shown in Fig.~\ref{fig:sample}.

\textbf{Part Mask Loss} $\mathcal{L}_{msk}$ provides semantic information for the location of body part:
\begin{equation}\small
\mathcal{L}_{msk}\!=\!\sum\nolimits_{k} (1 - \mathbf{IoU}(I_k, \hat{I}_k)),\; \mathbf{IoU}(I_k, \hat{I}_k)\!=\!\frac{|I_k \cap \hat{I}_k|}{|I_k\cup \hat{I}_k|},
\end{equation}
where $k$ is body part index and $\mathbf{IoU}(\cdot,\cdot)$ represents intersection over union of two masks.
We keep the same body segments (12 parts) $I$ and $(U, V)$ mapping as specified in~\cite{DensePose2018}.

\textbf{Adversarial Loss} $\mathcal{L}_{adv}$ constrains configuration plausibility. Unlike~\cite{kanazawa2018hmr} using unpaired or Mosh-based~\cite{Mosh} weakly-supervised SMPL annotations, we use ground-truth 3D human poses and body shapes from our synthetic dataset, which contains much larger action variations than most Mocap sequences (see Sec.~\ref{sec:dataset}). We believe such long-tail poses are crucial for the adversarial loss in finding the decision boundary. Hereby, we account for millions of synthetic samples as both paired ground truth and unpaired adversarial prior for realistic datasets. We follow the GAN loss definitions in~\cite{gan14} and train our generator and discriminator jointly.

\noindent {\small \textbullet} \, \textbf{3D Reconstruction Loss} $\mathcal{L}_{rec}$ measures the deformation of reconstructed 3D human body, compared with ground truth: 
\begin{equation}\small
\mathcal{L}_{rec} = \sum\nolimits_{i}^{K} \|P_i - \hat{P}_i \|_2,
\end{equation}
where $P_i$ and $\hat{P}_i$ represent 3D keypoint positions from input and generated 3D mesh, respectively.

\noindent {\small \textbullet} \, \textbf{Parameter Regression Loss} $\mathcal{L}_{rgr}$ measures mean square errors between estimated parameters $[\theta, \beta, \alpha]$ and ground truth $[\hat{\theta}, \hat{\beta}, \hat{\alpha}]$:
\begin{equation}\small
\mathcal{L}_{rgr} = \|[R_\theta, \beta, \alpha] - [R_{\hat{\theta}}, \hat{\beta}, \hat{\alpha}]\|_2,
\end{equation}
where $R_{\theta}$ denotes the rotation matrix of $\theta$. Notably, pose parameters are first transformed in rotation matrices. Losses are computed upon such matrices and gradients are automatically back-propagated. This helps avoid the singularity problem of XYZ-Euler based 3D rotation and requires no extra constraints on the rotation matrix, that is, 
\begin{equation}\small
R R^T = diag(1,\dots,1),\;\; \mathbf{det}(R) = 1,
\end{equation}
where $\mathbf{det}(\cdot)$ denotes the matrix determinant.

\beforesubsection
\subsection{Human Body Model} \label{sec:bodymodel}
\aftersubsection

We use a body shape model similar to SMPL~\cite{loper2015smpl,bogo2016keep}. The statistical body model is obtained by PCA on pose-normalized 3D models of real humans, obtained by non-rigid registration of a body template to 3D scans of the CAESAR dataset\footnote{\url{http://store.sae.org/caesar/}}, which represents anthropometric variability of 4,400 men and women.
The body template mesh has 7,324 vertices, 14,644 triangular faces and a skeletal rig with body and hand joints.

Our model is trained with all 3D scans in the dataset, resulting in a statistical model that can describe bodies from unseen in-the-wild images regardless of gender.
An arbitrary body shape can then be described by a set of shape coefficients (\ie, shape parameters or shape blend shapes) using a linear representation.
Truncating shape coefficients to 50 principal components enables reconstruction of all-gender body shapes without noticeable distortions:
\eg, the SMPL-Male with 10 coefficients does not reconstruct well female shape (RMSE=9.9mm), while an all-gender model does ({RMSE=6.3/3.4mm} with 10/50 coeffs respectively).

Considering potential applications in AR/VR, 3D animations and better utilization of annotations, we enrich the standard SMPL 24-joint skeleton with 28 joints for modeling fingers and 5 more joints on spine and head for better flexibility. We further add a root node for global translation and rotation, leading to a skeleton with 58 joints.

\beforesection
\section{MOCA Synthetic Dataset} \label{sec:dataset} 
\aftersection

The literature has provided several datasets to evaluate human 3D pose (\eg, H3.6M~\cite{ionescu2014human3}, MPI-INF-3DHP~\cite{Mehta3DV17}), but only few for joint 3D pose and body shape (\eg, SURREAL~\cite{varol2017learning} and UP-3D~\cite{lassner2017unite}).
However, SURREAL is dedicated to body segmentation and depth estimation and only has a rough skeleton (24 major body joints), while UP-3D has \emph{weakly-supervised} shapes (from SMPL fitted to LSP and MPII), arguably imprecise~\cite{varol18_bodynet}.

Hence, we propose \textbf{MOCA}
, a large-scale synthetic dataset with 2,089,104 images containing ground-truth body shapes and 3D poses, as shown in Fig.~\ref{fig:moca}. For various human poses and actions, we seek to a popular collection center of 3D human animations (\ie, Mixamo\footnote{\url{http://www.mixamo.com}}), whose sources mainly come from Mocap systems and artist designs. We implement a web crawler to fetch high fidelity animations. Notably, Mixamo supports tuning parameters (\eg, limb length, energy, overdrive) for each action sequence to generate variants. As we observe certain parameter settings may introduce artifacts, we thus keep the default setting for all sequences. We collect a set of 2,446 3D animation sequences with 261,138 frames at 30 fps, covering wide action categories of sports, combat, daily and social activities. We extract a finer 3D skeleton with fingers and facial bones using Maya and re-map those joints onto our body model.

We then generate 2,781 bodies using the 3D scans from CAESAR dataset and compute corresponding (PCA) shape coefficients. By combining 3D pose $\theta$ and shape $\beta$, we pose body models to specific pose\&shape configurations by standard linear blend skinning.

The complete combination of all 3D poses and body shapes produces an enormous amount of 3D human body samples. Currently, we randomly select 8 body shapes for each action sequence.
We further add a random camera view for each sequence, and render them as IUV image sequences using our IUV rasterizer (see Sec.~\ref{sec:randc}), obtaining a dataset with 2,089,104 frames in total and fully paired ground truth of body shape, 3D pose and the camera view.
For training/testing set partition, we set the ratio as 90\%/10\%. We synthesize the training set with the first 2,201 Mixamo action sequences and 2,502 CAESAR body shapes and leave the rest 246 action sequences and 279 body shapes only visible to the testing set.

\beforesection
\section{Experiments} \label{sec:experiments}
\aftersection

We evaluate DenseRaC on several public large-scale benchmarks for three tasks: 3D pose estimation, body shape estimation and body semantic segmentation.
We further assess human 3D reconstruction results (\ie, mesh-level reconstruction, joint\&shape parameter estimation) on the proposed large synthetic dataset MOCA that contains ground-truth 3D pose and body shape.
Our experiments compare favorably to the state of the art.
Estimated 3D poses and body shapes are stable on videos (see additional materials).
Our qualitative results also show natural hand poses (\eg, opened, clenched).

\beforesubsection
\subsection{Datasets}
\aftersubsection

\begin{figure}[ptb]
\centering
\includegraphics[width=\linewidth]{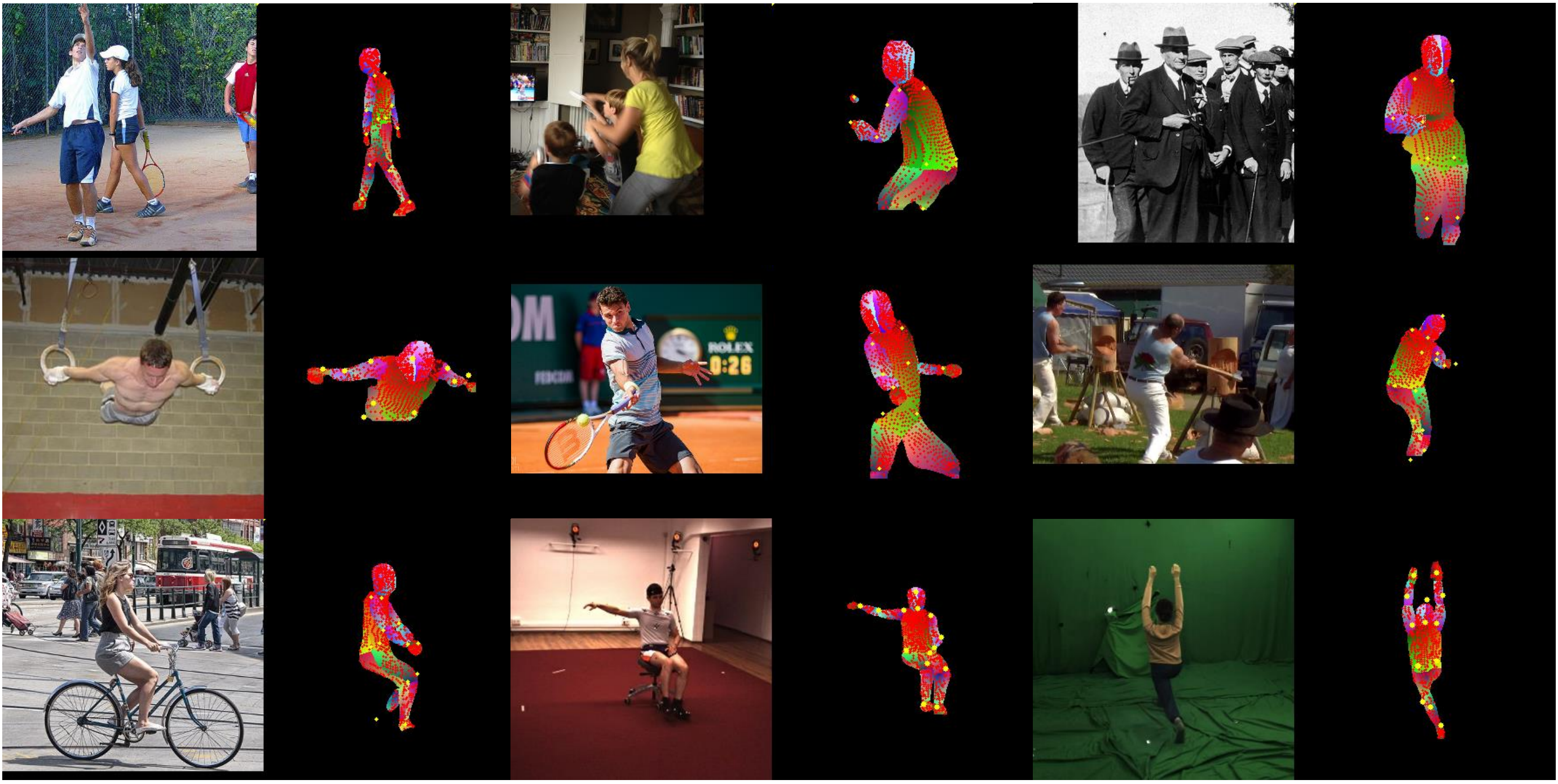}
\beforefigcaption
\caption{Pre-processed training samples from public benchmarks. Left: original image, right: estimated IUV image, ground-truth keypoint annotations (yellow) and dense landmarks (red).}
\afterfigcaption
\label{fig:sample}
\end{figure}

We use five public human benchmarks plus our synthesized MOCA for model training and evaluation, \ie, LSP~\cite{LSPDataset}, MPII~\cite{MPIIDataset}, COCO~\cite{COCODataset}, H3.6M~\cite{ionescu2011human3,ionescu2014human3} and MPI-INF-3DHP~\cite{Mehta3DV17}. We adopt standard training/validation/testing partitions on all datasets and calibrate loss terms using cross-validation. When a certain dataset is used for evaluation, all data from other datasets will be used in training.

For all training and testing samples, we crop out each person in the whole image using ground-truth bounding boxes. All samples are resized to $\sim$150-180 pixel height with preserved aspect ratio, and further adjusted to 224 $\times$ 224 with padding/cropping respectively. We then run IUV image estimation~\cite{DensePose2018} on all samples. Considering a sample $\mathcal{I}$ may contain multiple people and false alarms, we compute a saliency score $s = \frac{|m|}{\|m_c- \mathcal{I}_c\|_2}$ for each detected person mask $m$, where $m_c$ and $\mathcal{I}_c$ represent the center of the person mask and the image, respectively. We then pick the person mask with the largest saliency score and suppress the other detection responses. 

For the training set, we further run pixel-to-surface matching (as described in Sec.~\ref{sec:randc}) to create dense correspondences. We discard samples with less than 200 corresponding pairs, as IUV image estimation usually failed under such situation. As illustrated in Fig.~\ref{fig:sample}, pre-processing suppresses nuisances in the training samples quite well. During training, all training samples will further be augmented with a random jittering of translation, scaling and reflection to improve the model robustness. We also randomly black out a rectangle image region for the synthetic samples to simulate occlusion in realistic scenarios.

To unite the skeleton structure across all datasets, we use the same 14 joints as in LSP for joint related computation while maintaining our 58-joint skeleton in the backend. 

\beforesubsection
\subsection{Implementation Details}
\aftersubsection

In these experiments, the whole framework is implemented with TensorFlow and runs on a DGX workstation with 2 Intel E5 CPUs, 512GB memory and 8 Titan V100 GPUs.
Data synthesis and pre-processing (\ie, IUV image estimation) are implemented with multi-gpu data parallelism. The multi-gpu renderer processes around 300 fps and takes 2 days to generate 2 million MOCA samples (total size 2.7TB). Data pre-processing on realistic datasets takes 12 hours to prepare 0.8 million samples.

For learning, only a single GPU is used due to difficulty in gradient transfer and a potential performance drop. We use batch size 128, learning rate $10^{-5}$ for the generator and $10^{-4}$ for the discriminator, and Adam as the optimizer. Our full model is jointly trained on all datasets for 30 epochs. Empirically, for one batch, the forward pass takes around 15ms and the backward propagation takes ($\sim$130ms) with IUV image render-and-compare ($\sim$55ms) as the overhead. The total training process takes around a week to complete. For inference, IUV images are first estimated at around 15 fps and then the forward pass of our model is called, taking 120 fps and thus achieves \textit{real time}.

\beforesubsection
\subsection{3D Pose Estimation}
\aftersubsection

\begin{table}[ptb]
\centering
\setlength{\tabcolsep}{2pt}
\renewcommand\arraystretch{1.1}
\resizebox{\linewidth}{!}{
\begin{tabular}{@{}r|c||c||c@{}}
\hline\thickhline
\multirow{2}{*}{\textbf{H3.6M}} & Protocol \#1 & Protocol \#2 & Protocol \#3\\
\cline{2-4}
& MPJPE & MPJPE & MPJPE\\
\hline
Martinez \etal (ICCV'17)~\cite{martinez2017simple} & 62.9 & 47.7 & 84.8\\
Fang \etal (AAAI'18)~\cite{Fang2018} & 60.3 & 45.7 & \bf{72.8}\\
Rhodin \etal (CVPR'18)~\cite{helgeCVPR18mv3d} & 66.8 & - & -\\
Yang \etal (CVPR'18)~\cite{yang20183dposegan} & 58.6 & \bf{37.7} & - \\
Hossain \etal (ECCV'18)~\cite{mirECCV18temp3dpose} & \bf{51.9} & 42.0 & -\\
\thickhline
Lassner \etal (CVPR'17)~\cite{lassner2017unite} & 80.7 & - & -\\
HMR (CVPR'18)~\cite{kanazawa2018hmr} & 88.0 & 56.8 & 77.3\\
Pavlakos \etal (CVPR'18)~\cite{pavlakos2018ps} & - & 75.9 & -\\
NBF (3DV'18)~\cite{NeuralBodyFit18} & - & 59.9 & -\\
\hline
DenseRaC baseline         & 82.4 & 53.9 & 77.0 \\
+ render-and-compare & 79.5 & 51.4 & 75.9 \\
+ synthetic data     & 76.8 & 48.0 & 74.1 \\
\thickhline
\end{tabular}}
\resizebox{\linewidth}{!}{
\begin{tabular}{@{}r|c|c|c||c|c|c@{}}
\hline\thickhline
\multirow{2}{*}{\textbf{MPI-INF-3DHP}} & \multicolumn{3}{c||}{Protocol \#1}  & \multicolumn{3}{c}{Protocol \#2}\\
\cline{2-7}
& PCK & AUC & MPJPE & PCK & AUC & MPJPE \\
\hline
Mehta \etal (3DV'17)~\cite{Mehta3DV17} & 75.7 & 39.3 & 117.6 & - & - & - \\
Mehta \etal (TOG'17)~\cite{VnectToG17} & 76.6 & 40.4 & 124.7 & 83.9 & 47.3 & 98.0 \\
HMR (CVPR'18)~\cite{kanazawa2018hmr} & 72.9 & 36.5 & 124.2 & 86.3 & 47.8 & 89.8 \\
\hline
DenseRaC baseline & 73.1 & 36.7 & 123.1 & 86.8 & 47.8 & 88.7 \\
+ render-and-compare & 74.7 & 38.6 & 124.9 & 87.5 & 48.3 & 86.7 \\
+ synthetic data & \bf{76.9} & \bf{41.1} & \bf{114.2} & \bf{89.0} & \bf{49.1} & \bf{83.5} \\
\hline\thickhline
\end{tabular}}
\beforetab
\caption{Quantitative comparisons of mean per joint position error (MPJPE), PCK and AUC between the estimated 3D pose and ground truth on
\textbf{H3.6M} under \textit{Protocol \#1, \#2, \#3} and \textbf{MPI-INF-3DHP} under \textit{Protocol \#1, \#2}. - indicates results not reported. Lower MPJPE, higher PCK and AUC indicate better performance. Best scores are marked in \textbf{bold}.}
\aftertab
\label{tab:pose}
\end{table}

\begin{figure*}[ptb]
\begin{center}
\includegraphics[width=\textwidth]{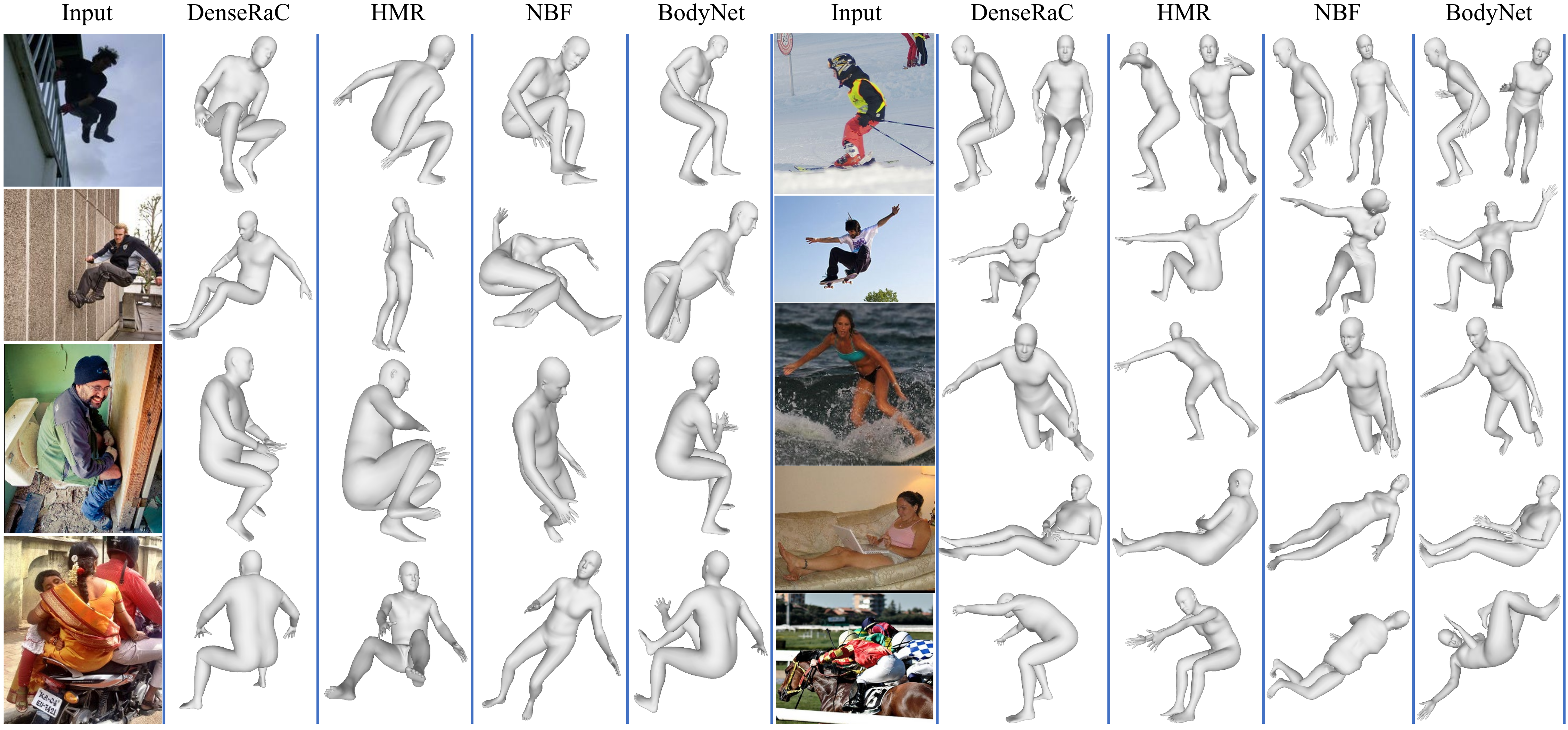}
\beforefigcaption
\caption{Qualitative comparisons of results estimated from DenseRaC versus state of the art~\cite{kanazawa2018hmr,NeuralBodyFit18,varol18_bodynet}. DenseRaC estimates 3D human poses and body shapes closest to the reality. Note that all examples come from the test set. Best viewed in color.}
\afterfigcaption
\label{fig:result_table}
\end{center}
\end{figure*}

We first evaluate our method for the task of 3D pose estimation on \textit{H3.6M}~\cite{ionescu2014human3} and \textit{MPI-INF-3DHP}~\cite{Mehta3DV17} datasets. 

For \textit{H3.6M}, we use three evaluation protocols used to measure the performance: i) \textbf{Protocol \#1} uses 5 subjects (S1, S5, S6, S7 and S8) for training and 2 subjects (S9 and S11) for testing. Sequences are down-sampled to 10 fps and all 4 cameras and trials are used for evaluation. MSE is measured between estimated and ground-truth 3D joints. ii) \textbf{Protocol \#2} selects the same subjects for training and testing as Protocol \#1, while evaluation is only conducted on sequences captured from the frontal camera (\ie, ``cam 3'') from trial 1 on all frames. Predictions are post-processed via rigid transformations (\ie, per-frame Procrustes analysis) before comparison. iii) \textbf{Protocol \#3} uses the same subjects, frame rates and trials for training and testing in Protocol \#1 except that camera views are further partitioned. The first three cameras (\ie, ``cam 0, 1, 2'') are used for training and the last camera (\ie, ``cam 3'') for testing.

\begin{table}[ptb]
\begin{center}
\setlength{\tabcolsep}{3mm}
\renewcommand\arraystretch{1.1}
\resizebox{\linewidth}{!}{
\begin{tabular}{@{}r|c|c||c|c@{}}
\hline\thickhline
\multirow{2}{*}{\textbf{UP-3D}} & \multicolumn{2}{c||}{Body Part}  & \multicolumn{2}{c}{Fg/Bg}\\
\cline{2-5}
& Accuracy & F1 & Accuracy & F1 \\
\hline
SMPL on DpCut (ECCV'16)~\cite{bogo2016keep} & 87.7 & 0.64 & 91.9 & \bf{0.88}\\
SMPL, UP-P91 (ICCV'17)~\cite{lassner2017unite} & 87.3 & 0.61 & 91.0 & 0.86 \\
HMR (CVPR'18)~\cite{kanazawa2018hmr} & 87.1 & 0.60 & 91.7 & 0.87 \\
BodyNet (ECCV'18)~\cite{varol18_bodynet} & - & - & \bf{92.8} & 0.84 \\
DenseRaC & \bf{87.9} & \bf{0.64} & 92.4 & \bf{0.88} \\
\hline\thickhline
\multirow{2}{*}{\textbf{MOCA}} & \multicolumn{2}{c||}{Body Part}  & \multicolumn{2}{c}{Fg/Bg}\\
\cline{2-5}
& Accuracy & F1 & Accuracy & F1 \\
\hline
HMR (CVPR'18)~\cite{kanazawa2018hmr} & 86.6 & 0.19 & 92.1 & 0.60 \\
DenseRaC & \bf{89.3} & \bf{0.27} & \bf{96.4} & \bf{0.68} \\
\hline\thickhline
\end{tabular}}
\beforetab
\caption{Quantitative comparisons of foreground and part segmentation on \textbf{UP-3D} and \textbf{MOCA} datasets. Accuracy unit is in \%. - indicates results not reported. Best scores are marked in \textbf{bold}.}
\aftertab
\end{center}
\label{tab:seg}
\end{table}

For \textit{MPI-INF-3DHP}, we use all sequences from S1-S7 as training set and sequences from S8 as testing set. We regard \textbf{Protocol \#1} as the default comparison and \textbf{Protocol \#2} as applying rigid transformations before comparison.

We compare our method with both task-oriented 3D pose state of the art~\cite{Sun2017,zhou2017towards,martinez2017simple,Mehta3DV17,VnectToG17,Fang2018,helgeCVPR18mv3d,yang20183dposegan,mirECCV18temp3dpose} and four parametric body model based estimators~\cite{lassner2017unite,kanazawa2018hmr,pavlakos2018ps,NeuralBodyFit18}. We set up two baselines to validate the effectiveness of two key components in the proposed framework: render-and-compare and joint learning with synthetic data. In ``DenseRaC baseline'', we use SMPL model and the same losses as ~\cite{kanazawa2018hmr}, only switch input sources from RGB images to IUV images. Variant ``+ render-and-compare'' denotes adding the proposed dense render-and-compare scheme losses into the framework and part masks. Variant ``+ synthetic data'' switches to our human body model and further uses augmented synthetic data for joint learning.

As reported in Table~\ref{tab:pose}, we can observe each component in DenseRaC contributes to the final performance
and leads DenseRaC to outperform state-of-the-art parametric body model estimators by a large margin.
Also notice DenseRaC is comparable with latest task-oriented 3D pose estimators.

\beforesubsection
\subsection{Human Body Segmentation}
\aftersubsection

Given rendered images from outputs, we further employ semantic segmentation as another task to measure how similar the reconstructed 3D human body looks to the person in the input image. We evaluate the tasks of human body segmentation and test our approach on the LSP subset of
UP-3D~\cite{lassner2017unite} and MOCA datasets. For UP-3D, we post-process our 24 body part masks by merging into the annotated 6 body part masks (i.e., head, torso, left and right leg, and left and right arm) and evaluate on body part and foreground segmentation, while we evaluate both body part segmentation (ignoring 4 subtle body parts, \ie, hands and feet) and foreground segmentation on MOCA. We measure segmentation accuracy and mean F1 score of the results and report metrics and comparisons in Table~\ref{tab:seg}.
It can be observed that our method achieves comparable or better performance with state of the art~\cite{bogo2016keep,lassner2017unite,kanazawa2018hmr,varol18_bodynet} on all datasets.

\beforesubsection
\subsection{3D Human Body Reconstruction}
\aftersubsection

\begin{table}[ptb]
\centering
\renewcommand\arraystretch{1.1}
\setlength{\tabcolsep}{3mm}
\resizebox{\linewidth}{!}{
\begin{tabular}{@{}l|c||c||c@{}}
\hline\thickhline
\multicolumn{1}{c|}{Methods} & MPJPE & MPVPE & $\text{MSE}_{\theta,\beta}$\\
\hline
HMR (CVPR'18)~\cite{kanazawa2018hmr} unpaired & 110.2 & - & - \\
HMR (CVPR'18)~\cite{kanazawa2018hmr} paired & 91.9 & - & - \\
\hline
DenseRaC, $\mathcal{L}_{rpj}^J$                                                                   & 133.0 & 174.5 & 18.227 \\
DenseRaC, $\mathcal{L}_{rpj}^J + \mathcal{L}_{adv}$                                               & 131.5 & 173.6 & 17.820 \\
DenseRaC, $\mathcal{L}_{rpj}^J + \mathcal{L}_{adv} + \mathcal{L}_{msk}$                           & 122.8 & 161.5 & 16.305 \\
DenseRaC, $\mathcal{L}_{rpj} + \mathcal{L}_{adv} + \mathcal{L}_{msk}$                             & 107.9 & 142.3 & 13.608 \\
DenseRaC, $\mathcal{L}_{rpj}^J + \mathcal{L}_{adv} + \mathcal{L}_{rec}^J$                         & 88.6 & 121.1 & 11.901 \\
DenseRaC, $\mathcal{L}_{rpj}^J + \mathcal{L}_{adv} + \mathcal{L}_{msk} + \mathcal{L}_{rec}^J$     & 86.5 & 119.8 & 10.496 \\
DenseRaC, $\mathcal{L}_{rpj} + \mathcal{L}_{adv} + \mathcal{L}_{rec}$                             & 82.9 & 111.0 & 8.943 \\
DenseRaC, $\mathcal{L}_{rpj} + \mathcal{L}_{adv} + \mathcal{L}_{msk} + \mathcal{L}_{rec}$         & 82.4 & 110.7 & 8.722 \\
DenseRaC, $\mathcal{L}_{rpj} + \mathcal{L}_{msk} + \mathcal{L}_{rec} + \mathcal{L}_{rgr}$         & 80.4 & 105.4 & 8.164 \\
DenseRaC, full                                                                                    & \bf{80.3} & \bf{105.2} & \bf{8.151} \\
\hline\thickhline
\end{tabular}}
\beforetab
\caption{Quantitative comparisons of MPJPE, MPVPE, Pose\&Shape Parameter Mean Square Error \textit{MSE}$_{\theta,\beta}$ on \textit{MOCA} dataset. Lower values are better. See text for detailed explanations.}
\aftertab
\label{tab:moca}
\end{table}

Notice 3D pose estimation and body semantic segmentation are tasks focusing on evaluating partial knowledge of the reconstructed 3D human body,
We further evaluate the reconstructed 3D human body using two metrics: Mean Per Mesh Vertex Position Error (MPVPE) and regression error on MOCA dataset.
These two metrics consider the 3D human body as a whole and provide more guidance about how well the reconstructed 3D human body is.
For comparison, we re-train HMR which takes IUV images as input and uses 2D/3D joint supervisions (\ie, only 14 2D/3D joints in LSP format) and their original unpaired data (Mosh~\cite{Mosh} on H3.6M and external Mocap) for the adversarial prior. As reported in Table~\ref{tab:moca}, DenseRaC still significantly outperforms the competitive method.

\textbf{Ablative Studies}. We set up variants of DenseRaC to validate effectiveness of each loss terms. We also define two loss variants $\mathcal{L}_{rpj}^J$ and $\mathcal{L}_{rec}^J$ representing 14-joint-only keypoint reprojection and 3D reconstruction losses, respectively. From the results, we could reach the following conclusions: i) All loss terms contribute to the final performance; ii) Losses used for dense render-and-compare provide richer information than those from sparse joints, greatly reduce impossible 3D body configurations; iii) When task oriented loss terms are given (\ie, $\mathcal{L}_{rec}$ and $\mathcal{L}_{rgr}$), the contribution from the dense render-and-compare scheme seems to be suppressed, yet such finer supervisions help DenseRaC reach a much better local optimum. 

\begin{figure}[ptb]
\centering
\includegraphics[width=\linewidth]{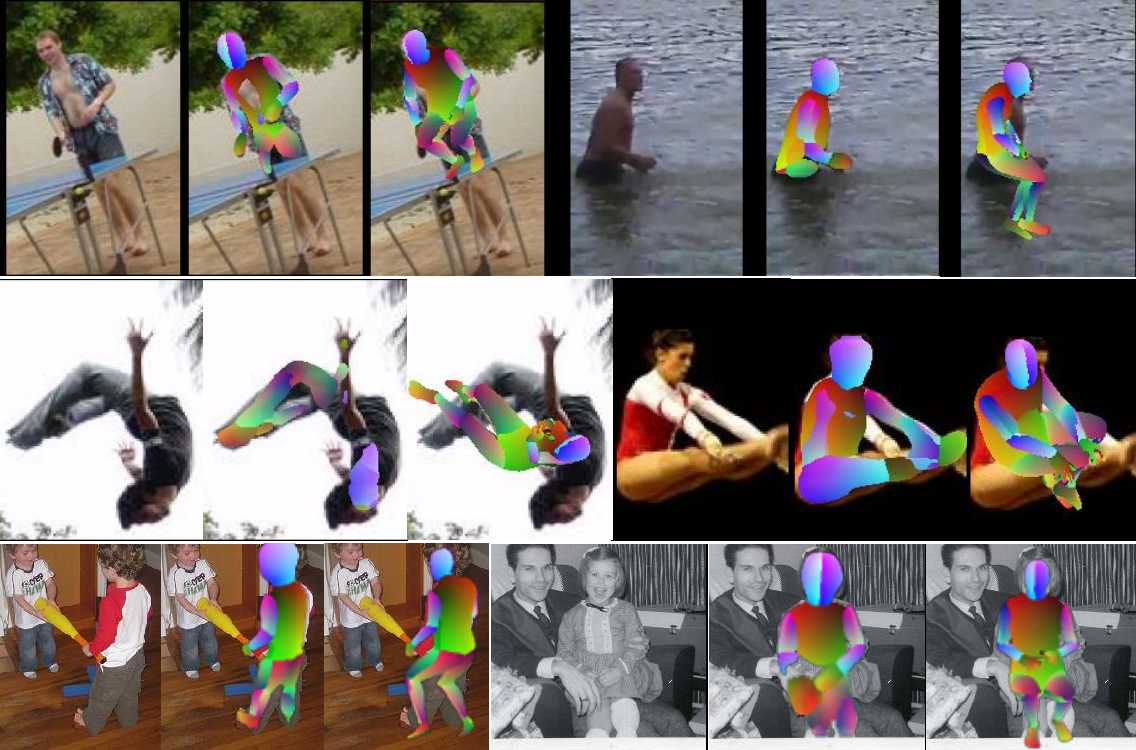}
\beforefigcaption
\caption{Current limitations: heavy occlusions (first row), incorrect IUV estimations (second row) and under-represented body shapes like children (third row). Each triplet shows the original image, IUV from~\cite{DensePose2018} (our model input), and our model output.}
\afterfigcaption
\label{fig:failcase}
\end{figure}

\textbf{Empirical Studies}. We present qualitative results and comparisons to have a better understanding of merits of our method.
As shown in Fig.~\ref{fig:result_table}, DenseRaC outperforms other competitive methods and reconstructs more plausible and natural 3D human bodies. Notably,
HMR, which relies on sparse landmarks, sometimes reconstructs plausible 3D human body appearance, but confuses body front and back.
Both NBF and BodyNet are sensitive to occlusions and heavy clothing. When fitting SMPL to such erroneously reconstructed volumes, BodyNet tends to produce highly non-human body shapes\footnote{We uses results from 3D skeleton fitting for BodyNet, as volume fitting usually performs much worse.}. For all three methods, the estimated human bodies are arguably in an average body shape and insensitive to genders. 
We also search failure cases on validation set, as shown in Fig.~\ref{fig:failcase}. DenseRaC suffers from errors in IUV estimations (\eg, occlusions, long-tail data), and is limited by the orthographic projection assumption and SMPL-based human body representation.

\begin{figure}[ptb]
\centering
\includegraphics[width=\linewidth]{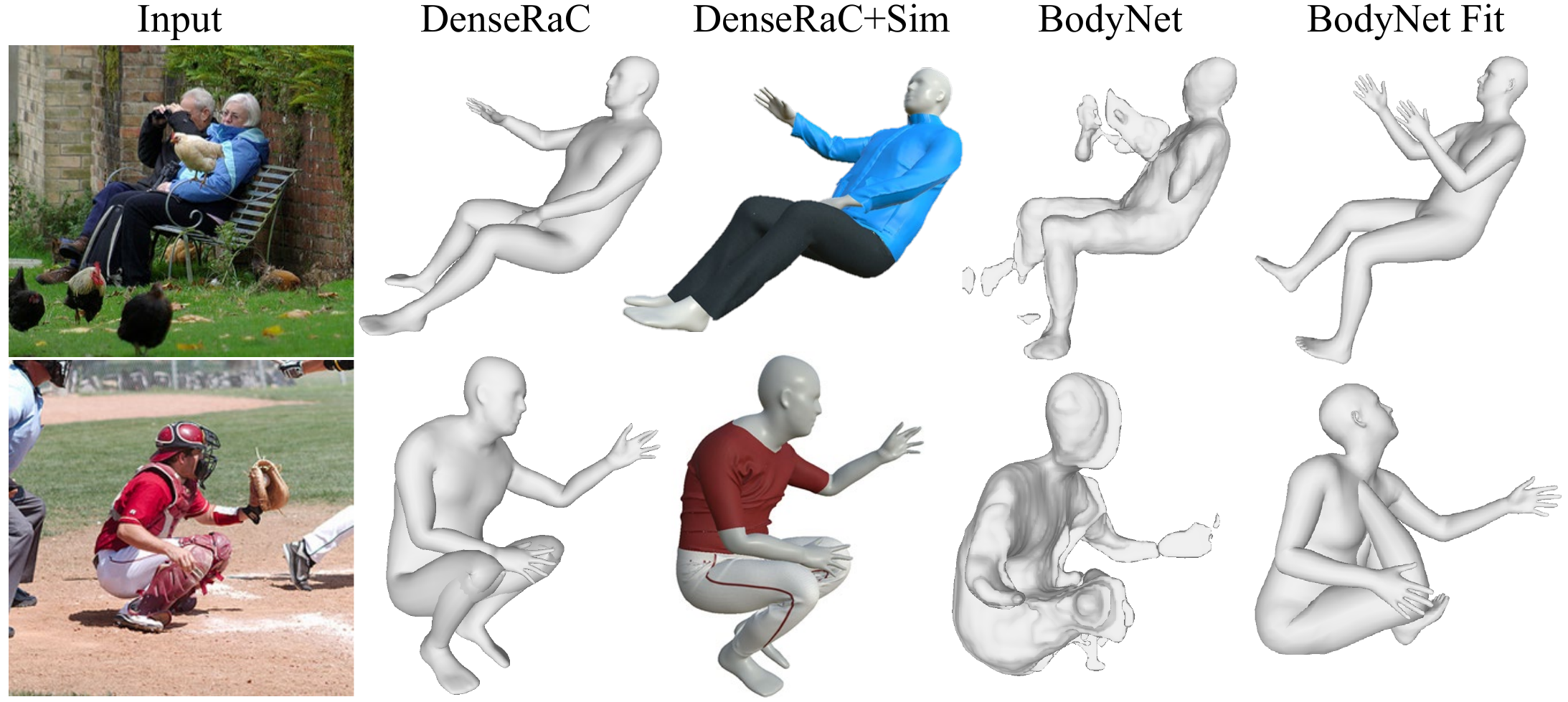}
\beforefigcaption
\caption{Comparisons for cascaded and end-to-end frameworks on the application of virtual dressing.}
\afterfigcaption
\label{fig:clothing}
\end{figure}

We also explored virtual dressing, namely draping virtual clothing on 3D human body, using our beneath-clothing estimation. As shown in Fig.~\ref{fig:intro} (top right) and Fig.~\ref{fig:clothing}, a cascaded framework for adding physical simulations of clothing is possible~\cite{guan2012drape,deepwrinkles18} and more visually acceptable than end-to-end volumetric reconstruction of BodyNet.

\beforesection
\section{Conclusion}
\aftersection

We propose DenseRaC, a new end-to-end framework for reconstructing 3D human body from monocular RGB images in the wild. DenseRaC utilizes the pixel-to-surface correspondence map as proxy representation and incorporates a dense render-and-compare scheme to minimize the gap between rendered outputs and inputs. We further boost the model training with large scale synthetic data (MOCA), mitigating the problem of unpaired training data. The proposed framework obtains superior performance and we will explore handling occlusion and interaction (e.g., by multi-view fusion~\cite{hangsceneparse18}, temporal smoothing~\cite{pavllovideopose3d19}) next.

\noindent {\small{\textbf{Acknowledgements.}} We would like to thank Tengyu Liu and Elan Markowitz for helping with data collection, Tuur Jan M Stuyck and Aaron Ferguson for cloth simulation, Natalia Neverova and colleagues at FRL, FAIR and UCLA for their support and advice.}

{\small
\bibliographystyle{ieee_fullname}
\bibliography{human_rec}

\begin{thebibliography}{10}\itemsep=-1pt

\bibitem{internetimages}
Images and videos available at \textit{youtube.com},
  \textit{onlinedoctor.superdrug.com}, \textit{shutterstock.com}.

\bibitem{agarwal2006recovering}
Ankur Agarwal and Bill Triggs.
\newblock Recovering {3D} human pose from monocular images.
\newblock {\em IEEE Transactions on Pattern Analysis and Machine Intelligence},
  28(1):44--58, 2006.

\bibitem{MPIIDataset}
Mykhaylo Andriluka, Leonid Pishchulin, Peter Gehler, and Bernt Schiele.
\newblock 2d human pose estimation: New benchmark and state of the art
  analysis.
\newblock In {\em IEEE Conference on Computer Vision and Pattern Recognition},
  2014.

\bibitem{bogo2016keep}
Federica Bogo, Angjoo Kanazawa, Christoph Lassner, Peter Gehler, Javier Romero,
  and Michael~J. Black.
\newblock Keep it {SMPL}: Automatic estimation of 3d human pose and shape from
  a single image.
\newblock In {\em European Conference on Computer Vision}, 2016.

\bibitem{cao2017realtime}
Zhe Cao, Tomas Simon, Shih-En Wei, and Yaser Sheikh.
\newblock Realtime multi-person 2d pose estimation using part affinity fields.
\newblock In {\em IEEE Conference on Computer Vision and Pattern Recognition},
  2017.

\bibitem{Dayan95AlySyn}
Peter Dayan, Geoffrey Hinton, Radford Neal, and Richard Zemel.
\newblock The helmholtz machine.
\newblock {\em Neural Computing}, 1995.

\bibitem{HSNetsl16}
Endri Dibra, Himanshu Jain, Cengiz Oztireli, Remo Ziegler, and Markus Gross.
\newblock Hs-nets: Estimating human body shape from silhouettes with
  convolutional neural networks.
\newblock In {\em International Conference on 3D Vision}, 2016.

\bibitem{fang2017rmpe}
Hao-Shu Fang, Shuqin Xie, Yu-Wing Tai, and Cewu Lu.
\newblock {RMPE}: Regional multi-person pose estimation.
\newblock In {\em IEEE International Conference on Computer Vision}, 2017.

\bibitem{Fang2018}
Hao-Shu Fang, Yuanlu Xu, Wenguan Wang, Xiaobai Liu, and Song-Chun Zhu.
\newblock Learning pose grammar to encode human body configuration for 3d pose
  estimation.
\newblock In {\em AAAI Conference on Artificial Intelligence}, 2018.

\bibitem{gan14}
Ian~J. Goodfellow, Jean Pouget-Abadie, Mehdi Mirza, Bing Xu, David
  Warde-Farley, Sherjil Ozair, Aaron Courville, and Yoshua Bengio.
\newblock Generative adversarial nets.
\newblock In {\em Annual Conference on Neural Information Processing Systems},
  2014.

\bibitem{guan2012drape}
Peng Guan, Loretta Reiss, David~A. Hirshberg, Alexander Weiss, and Michael~J.
  Black.
\newblock Drape: Dressing any person.
\newblock In {\em ACM SIGGRAPH}, 2012.

\bibitem{DensePose2018}
Riza~Alp Guler, Natalia Neverova, and Iasonas Kokkinos.
\newblock Densepose: Dense human pose estimation in the wild.
\newblock In {\em IEEE Conference on Computer Vision and Pattern Recognition},
  2018.

\bibitem{guler2017densereg}
Riza~Alp Guler, George Trigeorgis, Epameinondas Antonakos, Patrick Snape,
  Stefanos Zafeiriou, and Iasonas Kokkinos.
\newblock Densereg: Fully convolutional dense shape regression in-the-wild.
\newblock In {\em IEEE Conference on Computer Vision and Pattern Recognition},
  2017.

\bibitem{hattori2015learning}
Hironori Hattori, Vishnu Naresh~Boddeti, Kris~M Kitani, and Takeo Kanade.
\newblock Learning scene-specific pedestrian detectors without real data.
\newblock In {\em IEEE Conference on Computer Vision and Pattern Recognition},
  2015.

\bibitem{he2016deep}
Kaiming He, Xiangyu Zhang, Shaoqing Ren, and Jian Sun.
\newblock Deep residual learning for image recognition.
\newblock In {\em IEEE Conference on Computer Vision and Pattern Recognition},
  2016.

\bibitem{mirECCV18temp3dpose}
Mir Rayat~Imtiaz Hossain and James~J. Little.
\newblock Exploiting temporal information for 3d human pose estimation.
\newblock In {\em European Conference on Computer Vision}, 2018.

\bibitem{ionescu2011human3}
Catalin Ionescu, Fuxin Li, and Cristian Sminchisescu.
\newblock Latent structured models for human pose estimation.
\newblock In {\em IEEE International Conference on Computer Vision}, 2011.

\bibitem{ionescu2014human3}
Catalin Ionescu, Dragos Papava, Vlad Olaru, and Cristian Sminchisescu.
\newblock Human3. 6m: Large scale datasets and predictive methods for 3d human
  sensing in natural environments.
\newblock {\em IEEE Transactions on Pattern Analysis and Machine Intelligence},
  36(7):1325--1339, 2014.

\bibitem{umarECCV1825Dhtmp3dpose}
Umar Iqbal, Pavlo Molchanov, Thomas Breuel, Juergen Gall, and Jan Kautz.
\newblock Hand pose estimation via latent 2.5d heatmap regression.
\newblock In {\em European Conference on Computer Vision}, 2018.

\bibitem{ShapeFMask18Arxiv}
Zhongping Ji, Xiao Qi, Yigang Wang, Gang Xu, Peng Du, and Qing Wu.
\newblock Shape-from-mask: A deep learning based human body shape
  reconstruction from binary mask images.
\newblock {\em arXiv preprint arXiv:1806.08485}, 2018.

\bibitem{LSPDataset}
Sam Johnson and Mark Everingham.
\newblock Learning effective human pose estimation from inaccurate annotation.
\newblock In {\em IEEE Conference on Computer Vision and Pattern Recognition},
  2011.

\bibitem{kanazawa2018hmr}
Angjoo Kanazawa, Michael~J. Black, David~W Jacobs, and Jitendra Malik.
\newblock End-to-end recovery of human shape and pose.
\newblock In {\em IEEE Conference on Computer Vision and Pattern Recognition},
  2018.

\bibitem{kato2018renderer}
Hiroharu Kato, Yoshitaka Ushiku, and Tatsuya Harada.
\newblock Neural 3d mesh renderer.
\newblock In {\em IEEE Conference on Computer Vision and Pattern Recognition},
  2018.

\bibitem{Kundu18RenderCompare}
Abhijit Kundu, Yin Li, and James Rehg.
\newblock 3d-rcnn: Instance-level 3d object reconstruction via
  render-and-compare.
\newblock In {\em IEEE Conference on Computer Vision and Pattern Recognition},
  2018.

\bibitem{deepwrinkles18}
Zorah Laehner, Daniel Cremers, and Tony Tung.
\newblock Deepwrinkles: Accurate and realistic clothing modeling.
\newblock In {\em European Conference on Computer Vision}, 2018.

\bibitem{lassner2017unite}
Christoph Lassner, Javier Romero, Martin Kiefel, Federica Bogo, Michael~J.
  Black, and Peter~V Gehler.
\newblock Unite the people: Closing the loop between 3d and 2d human
  representations.
\newblock In {\em IEEE Conference on Computer Vision and Pattern Recognition},
  2017.

\bibitem{li2015maximum}
Sijin Li, Weichen Zhang, and Antoni~B Chan.
\newblock Maximum-margin structured learning with deep networks for 3d human
  pose estimation.
\newblock In {\em IEEE International Conference on Computer Vision}, 2015.

\bibitem{COCODataset}
Tsung-Yi Lin, Michael Maire, Serge Belongie, James Hays, Pietro Perona, Deva
  Ramanan, Piotr Dollar, and C.~Lawrence Zitnick.
\newblock Microsoft coco: Common objects in context.
\newblock In {\em European Conference on Computer Vision}, 2014.

\bibitem{Mosh}
Matthew Loper, Naureen Mahmood, and Michael Black.
\newblock Mosh: Motion and shape capture from sparse markers.
\newblock In {\em SIGGRAPH Asia}, 2014.

\bibitem{loper2015smpl}
Matthew Loper, Naureen Mahmood, Javier Romero, Gerard Pons-Moll, and Michael~J.
  Black.
\newblock Smpl: A skinned multi-person linear model.
\newblock {\em ACM Transactions on Graphics}, 34(6):248, 2015.

\bibitem{loper2014opendr}
Matthew~M. Loper and Michael~J. Black.
\newblock Opendr: An approximate differentiable renderer.
\newblock In {\em European Conference on Computer Vision}, 2014.

\bibitem{diogoCVPR182d3dml}
Diogo~C. Luvizon, David Picard, and Hedi Tabia.
\newblock 2d/3d pose estimation and action recognition using multitask deep
  learning.
\newblock In {\em IEEE Conference on Computer Vision and Pattern Recognition},
  2018.

\bibitem{marschner2015}
Steve Marschner and Peter Shirley.
\newblock Fundamentals of computer graphics.
\newblock In {\em CRC Press}, 2015.

\bibitem{martinez2017simple}
Julieta Martinez, Rayat Hossain, Javier Romero, and James~J Little.
\newblock A simple yet effective baseline for 3d human pose estimation.
\newblock In {\em IEEE International Conference on Computer Vision}, 2017.

\bibitem{3Dvideo12}
Takashi Matsuyama, Shohei Nobuhara, Takeshi Takai, and Tony Tung.
\newblock 3d video and its applications.
\newblock In {\em Springer}, 2012.

\bibitem{Mehta3DV17}
Dushyant Mehta, Helge Rhodin, Dan Casas, Pascal Fua, Oleksandr Sotnychenko,
  Weipeng Xu, and Christian Theobalt.
\newblock Monocular 3d human pose estimation in the wild using improved cnn
  supervision.
\newblock In {\em International Conference on 3D Vision}, 2017.

\bibitem{VnectToG17}
Dushyant Mehta, Srinath Sridhar, Oleksandr Sotnychenko, Helge Rhodin, Mohammad
  Shafiei, Hans-Peter Seidel, Weipeng Xu, Dan Casas, and Christian Theobalt.
\newblock Vnect: Real-time 3d human pose estimation with a single rgb camera.
\newblock In {\em ACM Transactions on Graphics}, 2017.

\bibitem{Nie2017}
Bruce~Xiaohan Nie, Ping Wei, and Song-Chun Zhu.
\newblock Monocular 3d human pose estimation by predicting depth on joints.
\newblock In {\em IEEE International Conference on Computer Vision}, 2017.

\bibitem{NeuralBodyFit18}
Mohamed Omran, Christoph Lassner, Gerard Pons-Moll, Peter~V. Gehler, and Bernt
  Schiele.
\newblock Neural body fitting: Unifying deep learning and model-based human
  pose and shape estimation.
\newblock In {\em International Conference on 3D Vision}, 2018.

\bibitem{paul2003fast}
Gregory~Shakhnarovich Paul, Paul Viola, and Trevor Darrell.
\newblock Fast pose estimation with parameter-sensitive hashing.
\newblock In {\em IEEE International Conference on Computer Vision}, 2003.

\bibitem{pavlakos2017volumetric}
Georgios Pavlakos, Xiaowei Zhou, Konstantinos~G Derpanis, and Kostas
  Daniilidis.
\newblock Coarse-to-fine volumetric prediction for single-image 3{D} human
  pose.
\newblock In {\em IEEE Conference on Computer Vision and Pattern Recognition},
  2017.

\bibitem{pavlakos2018ps}
Georgios Pavlakos, Luyang Zhu, Xiaowei Zhou, and Kostas Daniilidis.
\newblock Learning to estimate 3d human pose and shape from a single color
  image.
\newblock In {\em IEEE Conference on Computer Vision and Pattern Recognition},
  2018.

\bibitem{pavllovideopose3d19}
Dario Pavllo, Christoph Feichtenhofer, David Grangier, and Michael Auli.
\newblock 3d human pose estimation in video with temporal convolutions and
  semi-supervised training.
\newblock In {\em Conference on Computer Vision and Pattern Recognition
  (CVPR)}, 2019.

\bibitem{pishchulin2012articulated}
Leonid Pishchulin, Arjun Jain, Mykhaylo Andriluka, Thorsten Thormahlen, and
  Bernt Schiele.
\newblock Articulated people detection and pose estimation: Reshaping the
  future.
\newblock In {\em IEEE Conference on Computer Vision and Pattern Recognition},
  2012.

\bibitem{hangsceneparse18}
Hang Qi, Yuanlu Xu, Tao Yuan, Tianfu Wu, and Song-Chun Zhu.
\newblock Scene-centric joint parsing of cross-view videos.
\newblock In {\em AAAI Conference on Artificial Intelligence}, 2018.

\bibitem{rahmani20163d}
Hossein Rahmani and Ajmal Mian.
\newblock 3d action recognition from novel viewpoints.
\newblock In {\em IEEE Conference on Computer Vision and Pattern Recognition},
  2016.

\bibitem{helgeECCV18GeoAware}
Helge Rhodin, Mathieu Salzmann, and Pascal Fua.
\newblock Unsupervised geometry-aware representation for 3d human pose
  estimation.
\newblock In {\em European Conference on Computer Vision}, 2018.

\bibitem{helgeCVPR18mv3d}
Helge Rhodin, Jorg Sporri, Isinsu Katircioglu, Victor Constantin, Frederic
  Meyer, Erich Muller, Mathieu Salzmann, and Pascal Fua.
\newblock Learning monocular 3d human pose estimation from multi-view images.
\newblock In {\em IEEE Conference on Computer Vision and Pattern Recognition},
  2018.

\bibitem{rogez2016mocap}
Gr{\'e}gory Rogez and Cordelia Schmid.
\newblock Mocap-guided data augmentation for 3d pose estimation in the wild.
\newblock In {\em Annual Conference on Neural Information Processing Systems},
  2016.

\bibitem{Sun2017}
Xiao Sun, Jiaxiang Shang, Shuang Liang, and Yichen Wei.
\newblock Compositional human pose regression.
\newblock In {\em IEEE International Conference on Computer Vision}, 2017.

\bibitem{TanBC17}
Vince Tan, Ignas Budvytis, and Roberto Cipolla.
\newblock Indirect deep structured learning for 3d human body shape and pose
  prediction.
\newblock In {\em British Machine Vision Conference}, 2017.

\bibitem{tekin2016structured}
Bugra Tekin, Isinsu Katircioglu, Mathieu Salzmann, Vincent Lepetit, and Pascal
  Fua.
\newblock Structured prediction of {3D} human pose with deep neural networks.
\newblock In {\em British Machine Vision Conference}, 2016.

\bibitem{tung2017self}
Hsiao-Yu Tung, Hsiao-Wei Tung, Ersin Yumer, and Katerina Fragkiadaki.
\newblock Self-supervised learning of motion capture.
\newblock In {\em NIPS}, 2017.

\bibitem{tungICCV09}
Tony Tung, Shohei Nobuhara, and Takashi Matsuyama.
\newblock Complete multi-view reconstruction of dynamic scenes from
  probabilistic fusion of narrow and wide baseline stereo.
\newblock In {\em IEEE International Conference on Computer Vision}, 2009.

\bibitem{varol18_bodynet}
Gul Varol, Duygu Ceylan, Bryan Russell, Jimei Yang, Ersin Yumer, Ivan Laptev,
  and Cordelia Schmid.
\newblock Bodynet: Volumetric inference of 3d human body shapes.
\newblock In {\em European Conference on Computer Vision}, 2018.

\bibitem{varol2017learning}
Gul Varol, Javier Romero, Xavier Martin, Naureen Mahmood, Michael~J. Black,
  Ivan Laptev, and Cordelia Schmid.
\newblock Learning from synthetic humans.
\newblock In {\em IEEE Conference on Computer Vision and Pattern Recognition},
  2017.

\bibitem{XuFashionCVPR2018}
Wenguan Wang, Yuanlu Xu, Jianbing Shen, and Song-Chun Zhu.
\newblock Attentive fashion grammar network for fashion landmark detection and
  clothing category classification.
\newblock In {\em IEEE Conference on Computer Vision and Pattern Recognition},
  2018.

\bibitem{xu2013reid}
Yuanlu Xu, Liang Lin, Wei-Shi Zheng, and Xiaobai Liu.
\newblock Human re-identification by matching compositional template with
  cluster sampling.
\newblock In {\em IEEE International Conference on Computer Vision}, 2013.

\bibitem{xu2018caog}
Yuanlu Xu, Lei Qin, Xiaobai Liu, Jianwen Xie, and Song-Chun Zhu.
\newblock A causal and-or graph model for visibility fluent reasoning in
  tracking interacting objects.
\newblock In {\em IEEE Conference on Computer Vision and Pattern Recognition},
  2018.

\bibitem{yang20183dposegan}
Wei Yang, Wanli Ouyang, Xiaolong Wang, Jimmy Ren, Hongsheng Li, and Xiaogang
  Wang.
\newblock 3d human pose estimation in the wild by adversarial learning.
\newblock In {\em IEEE Conference on Computer Vision and Pattern Recognition},
  2018.

\bibitem{yasin2016dual}
Hashim Yasin, Umar Iqbal, Bjorn Kruger, Andreas Weber, and Juergen Gall.
\newblock A dual-source approach for 3d pose estimation from a single image.
\newblock In {\em IEEE Conference on Computer Vision and Pattern Recognition},
  2016.

\bibitem{zhou2017towards}
Xingyi Zhou, Qixing Huang, Xiao Sun, Xiangyang Xue, and Yichen Wei.
\newblock Towards {3D} human pose estimation in the wild: a weakly-supervised
  approach.
\newblock In {\em IEEE International Conference on Computer Vision}, 2017.

\bibitem{zhou2016sparseness}
Xiaowei Zhou, Menglong Zhu, Spyridon Leonardos, Konstantinos~G Derpanis, and
  Kostas Daniilidis.
\newblock Sparseness meets deepness: 3d human pose estimation from monocular
  video.
\newblock In {\em IEEE Conference on Computer Vision and Pattern Recognition},
  2016.

\bibitem{zuffi2015stitched}
Silvia Zuffi and Michael~J. Black.
\newblock The stitched puppet: A graphical model of 3d human shape and pose.
\newblock In {\em IEEE Conference on Computer Vision and Pattern Recognition},
  2015.

\end{thebibliography}
}

\end{document}